\documentclass{article} 
\usepackage{iclr2023_conference,times}

\usepackage{amsmath,amsfonts,bm}

\def\eqref#1{equation~\ref{#1}}

\def\1{\bm{1}}

\def\vk{{\bm{k}}}

\def\vq{{\bm{q}}}

\def\vv{{\bm{v}}}

\def\mA{{\bm{A}}}

\def\mE{{\bm{E}}}

\def\mK{{\bm{K}}}

\def\mO{{\bm{O}}}

\def\mQ{{\bm{Q}}}

\def\mV{{\bm{V}}}

\DeclareMathAlphabet{\mathsfit}{\encodingdefault}{\sfdefault}{m}{sl}
\SetMathAlphabet{\mathsfit}{bold}{\encodingdefault}{\sfdefault}{bx}{n}

\newcommand{\R}{\mathbb{R}}

\usepackage{hyperref}
\usepackage{url}
\usepackage[pdftex]{graphicx}

\usepackage{graphics}

\usepackage[usestackEOL]{stackengine}
\strutlongstacks{T}

\title{Grouped self-attention mechanism for a memory-efficient Transformer}

\author{Bumjun Jung$^1$, Yusuke Mukuta$^{1,2}$ \& Tatsuya Harada$^{1,2}$ \\
$^1$The University of Tokyo, Japan\\
$^2$RIKEN AIP, Japan \\
\texttt{\{jung,mukuta,harada\}@mi.t.u-tokyo.ac.jp} \\
}

\begin{document}

\maketitle

\begin{abstract}
Time-series data analysis is important because numerous real-world tasks such as forecasting weather, electricity consumption, and stock market involve predicting data that vary over time. Time-series data are generally recorded over a long period of observation with long sequences owing to their periodic characteristics and long-range dependencies over time. Thus, capturing long-range dependency is an important factor in time-series data forecasting. To solve these problems, we proposed two novel modules, Grouped Self-Attention (GSA) and Compressed Cross-Attention (CCA). With both modules, we achieved a computational space and time complexity of order $O(l)$ with a sequence length $l$ under small hyperparameter limitations, and can capture locality while considering global information. The results of experiments conducted on time-series datasets show that our proposed model efficiently exhibited reduced computational complexity and performance comparable to or better than existing methods.



\end{abstract}

\section{Introduction}
Time-series data forecasting is important because many real-world tasks are formatted as problems based on time series. For example, both weather forecasting and predicting electricity consumption involve time-series forecasting. These tasks usually require an output sequence of considerable length to perform prediction as well as a long input sequence to capture the long-range dependencies among data, which is an important factor in time-series forecasting.

The Transformer model (\cite{vaswani2017attention}) is among the most powerful deep learning architectures, and it has been shown to capture long-range dependencies very well compared to other DNN methods such as RNN or LSTM-based models (\cite{hua2019deep,yadav2020optimizing}) because it does not use a recurrent structure and can directly access and refer to past sequence information.  Therefore, applying Transformer models to time-series forecasting tasks may be considered a promising approach. Recent studies (\cite{zhou2021informer, wang2020linformer}) have sought to apply the Transformer architecture to time-series data forecasting to utilize its powerful performance. 


However, two important obstacles to directly applying the Transformer architecture to time-series forecasting remain to be resolved. First, the order of the computational and space complexity increases quadratically for a sequence of length $l$ in the self-attention module, which is the bottleneck of the Transformer architecture. This problem particularly affects time-series data application, which tend to require longer sequences. In the time-series domain, the required sequence length may easily exceed 500 to 1000 and may require sequences of massive length compared to other domains for some applications, such as voice or biosignal data with a sampling rate in units of kHz. For example, in the natural language domain, the maximum sequence length of BERT (\cite{devlin2018bert}), a frequently used pre-trained NLP model, is 512, and it can handle almost any input sentence. 

In addition, the computational complexity of a Transformer model is affected not only by the length of the input sequence but also by that of the prediction sequence. The cross-attention module provides encoded information of the input to the decoder layer and the complexity of the cross-attention module increases with the order of multiplication of the lengths of the input and output sequences. These problems often cause a memory shortage error on GPU servers, which imposes a limit on the extent to which the sequence length can be increased to obtain higher performance or longer predictions.

To summarize, The algorithm has an order of $O(l^2)$ computational time and space complexity in the self-attention module, where $l$ denotes the sequence length. Moreover, the complexity of the cross-attention module is $O(l_{enc}\times l_{dec})$ where $\l_{enc}$ and $ l_{dec}$ denote the input and output sequence lengths, respectively.

To address these problems, we propose Grouped Transformer as a memory-efficient, high-performance transformer. The Grouped Transformer architecture consists of two novel modules, including Grouped Self-Attention (GSA) and Compressed Cross-Attention (CCA) mechanisms. 

\begin{figure}[tbp]
 \centering
 \includegraphics[width=0.9\textwidth]{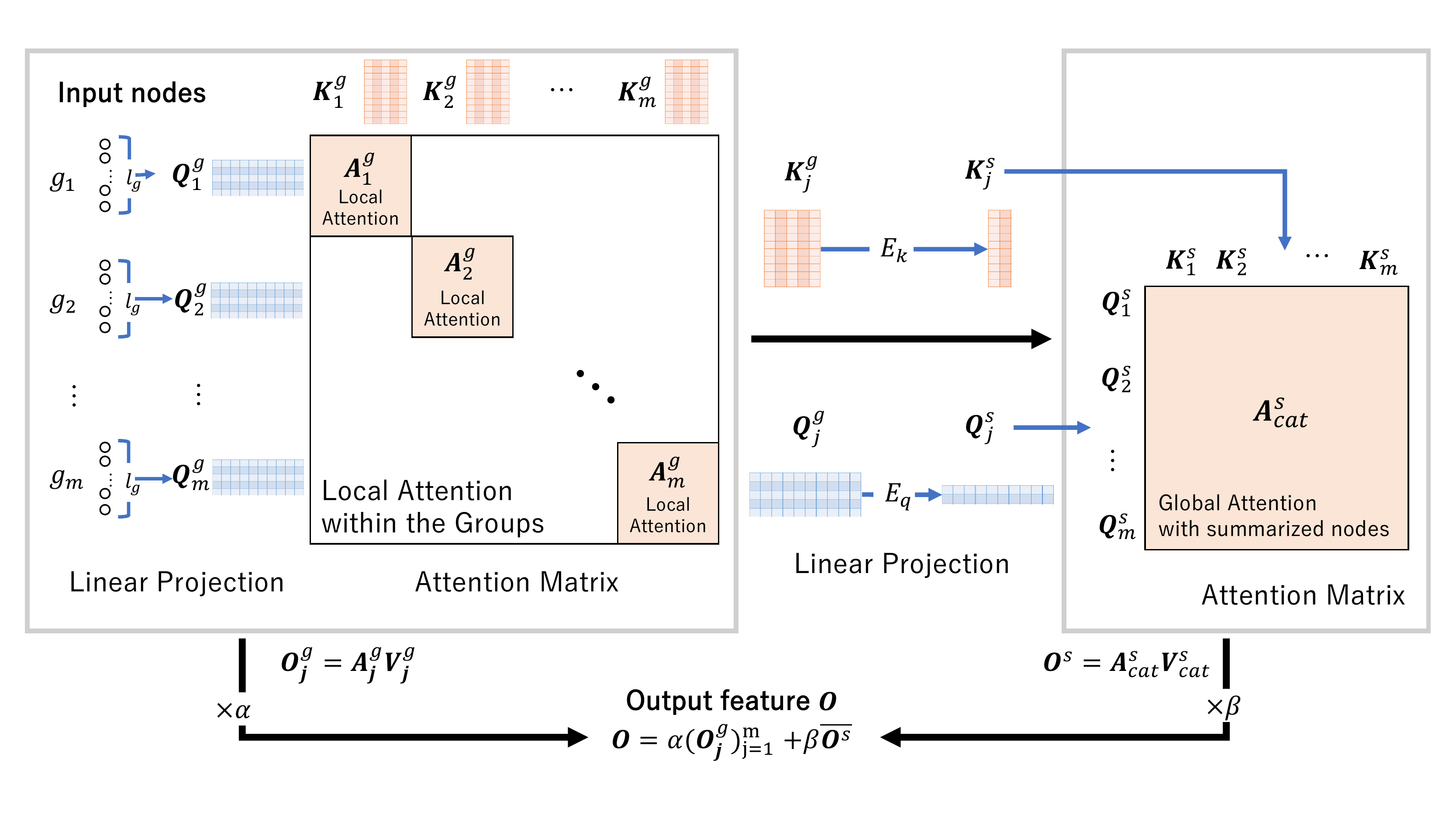}
 \caption[Grouped self-attention]{Overall illustration of the Grouped Self-Attention module.}
 \label{fig:grouped_self_attention}
\end{figure}

GSA module is a type of local attention mechanism that can also reflect global attention features. First, the input nodes are divided into several groups and attention is calculated only within that local block. For each group, the summarized queries, keys, and values are generated by linear layers to form summarized nodes that represents the summarized information of the nodes within the group. All summarized nodes from every group are concatenated and self-attention is calculated among those nodes to reflect global features from other groups as well. The outputs of local attention within the group and global attention output are merged to generate a final output which captures locality while also reflecting global information. Moreover, by limiting the number of summarized nodes included in the global attention calculation, the computational complexity can be reduced to order of $O(l)$.

CCA mechanism can reduce computational complexity while minimizing the loss of information. In contrast to the encoder distillation in Informer (\cite{zhou2021informer}), in which the encoder sequence length is halved through each encoder layer, CCA compresses the length of the encoder output to a fixed length in each decoder layer. By using different weights of linear layers in each decoder layer, the encoded features are compressed with various weights to minimize information loss during compression.

Our proposed method solves the issues mentions above by applying GSA and CCA modules. We conducted experiments with the time-series datasets provided by \cite{zhou2021informer}. The results were promising because the proposed method used less memory and required less computational time with performance comparable to or better than that of existing methods. Moreover, the results of experiments on the computational complexity of our proposed approach show that our method exhibited a complexity of order $O(l)$ for sequence length $l$. The results of an ablation study on the effects of GSA module also demonstrate that the proposed method provides a powerful learning architecture that can efficiently reflect the global features.

Fig.\ref{fig:grouped_self_attention} illustrates the overall mechanism of the GSA module and Fig.\ref{fig:grouped_self_attention_detailed} illustrates the computation held in GSA in detail.

The contributions of this study are summarized as follows.
\begin{itemize}
\item We propose a GSA module designed to efficiently capture locality information with global features while reducing the computational costs to linear order.
\item The CCA module successfully reduces the computational complexity of the proposed approach with less information loss with separated linear weights in each decoder layer.
\item The experiments demonstrated that our proposed method successfully managed to reduce the memory usage while achieving performance comparable to or better than that of the baseline methods. 
\end{itemize}

\section{Related Works}
Studies have considered the application of Transformer models to long sequence data. Existing methods to reduce the complexity of the Transformer model may be mainly divided into two categories, including sparse attention mechanisms and approximation methods, each of which involves some advantages and disadvantages.

This section introduces the mechanism of the self-attention module in Transformer in detail along with research focused on the quadratic complexity issue with self-attention.

\subsection{Transformer and the self-attention module}

Transformer (\cite{vaswani2017attention}) is a popular deep learning architecture utilized in many different applications. Their powerful performance on sequence data has been demonstrated in numerous domains. They are thus utilized across many domains such as natural language processing (NLP) and computer vision (CV). Many state-of-the-art methods for various tasks in such domains use a Transformer architecture, which is an encoder-decoder architecture based on self-attention and cross-attention modules. 

The core mechanism of Transformer is the self-attention module. The self-attention module utilizes scaled dot-product attention with an input comprising queries and keys, and values of dimension $d$ generated from each input node of the Transformer model. Scaled dot-product attention is calculated by dividing dot product results of queries and keys by $\sqrt{d}$ and applying a softmax function to obtain the weights of the values. To compute the attention function on a set of queries simultaneously, queries, keys, and values are concatenated into a matrix $\mQ, \mK, \mV$ where $\mQ\in \R^{l_Q\times d}$, $\mK\in \R^{l_K\times d}$, and $\mV\in \R^{l_K\times d}$ as given below. 

\begin{equation}
  \label{eq:attention_mat}
    \mA = softmax(\frac{\mQ\displaystyle \mK^{T}}{\sqrt{d}}),
\end{equation}

\begin{equation}
  \label{eq:self-attention}
    Attention(\mQ, \mK, \mV) = softmax(\frac{\mQ\mK^{T}}{\sqrt{d}})\mV = \mA\mV.
\end{equation}

However, the self-attention mechanism is also the bottleneck of the Transformer architecture owing to the quadratic growth of its computational complexity for the sequence length. This is caused by the calculation of the attention matrix where the dot-product operation is performed $l_Q\times l_K$ times and occupies memory in proportion to $l_Q\times l_K$ as well. 

\subsection{Efficient self-attention mechanisms}
Capturing long-range dependency is a crucial factor in time-series data analysis. Compressive Transformer (\cite{rae2019compressive}) and Transformer-XL (\cite{dai2019transformer}) reinforced the ability of capturing long-range dependency by utilizing auxiliary hidden states. However, these approaches could amplify the computational order issue mentioned above, especially with longer input sequences.

Recent studies have addressed this issue to apply the Transformer architecture to tasks involving a long sequence length by creating an efficient way to calculate attention values.

\subsubsection{Sparse attention mechanisms}
Research on sparse attention mechanisms such as Informer (\cite{zhou2021informer}), Sparse Transformer (\cite{child2019generating}), Longformer (\cite{beltagy2020longformer}), and LogSparse Transformer (\cite{li2019enhancing}) addressed the issue of complexity by calculating the attention matrix sparsely, focusing on the dominant attention values evaluated with their own heuristic metrics. Our proposed model is based on the Informer framework, which utilized global time stamp embedding and a generative style decoder, and a different self-attention module is applied. Moreover, Informer used encoder distillation to shrink the encoder output length by half for each encoder layer to reduce the computational complexity in the cross-attention module. However, the way they choose the nodes to calculate attention involved random sampling, which may ignore the calculation of dominant attention. Other methods such as Sparse Transformer, Longformer, and LogSparse Transformer require the premise that near nodes or nodes selected with certain rules have dominant attention values, which may not be compatible with data that correlates with distant nodes. 

Local attention is another example of sparse attention mechanisms that can reflect locality while efficiently calculating attention. However, accessing global information is difficult because distant nodes are not included in the calculation of local attention. Image Transformer (\cite{parmar2018image}) utilized local1d attention to generate image pixels with the premise that close nodes generated from near pixels hold important information. On the other hand, our proposed method not only focuses on locality with local attention but also reflects global information by calculating attention for summarized nodes from each group. Moreover, \cite{chu2021twins} introduces the locally-grouped self-attention which does grouped local attention with global attention to learn the image representations. However, the way of calculating local attention and global attention is different from the proposed method and it does not suppress the complexity order of the self-attention module for efficient computation.

\subsubsection{Approximation methods}
Researchers have introduced approximation methods in self-attention mechanisms to reduce the size of the attention matrix or the number of calculations required. Linformer (\cite{wang2020linformer}) utilized a low-rank approximation on the attention matrix and Performer (\cite{choromanski2020rethinking}) used a softmax kernel approximation to reorder the attention calculation and reduce computational complexity. Research that applied approximation methods successfully achieved a complexity of order $O(l)$ while reflecting global information. However, information is inevitably lost in approximation, and although this approach can capture global features, it is not guaranteed to preserve locality information.

\section{Proposed Method}
Our proposed model consists of two main contributions, including Grouped Self-Attention (GSA) and Compressed Cross-Attention (CCA) modules. 
The GSA module divides the nodes into multiple groups and constructs blocks of an attention matrix to calculate local attention. Each group’s nodes are projected to form summarized nodes, and the self-attention is calculated among those summarized nodes to reflect the global features. The outputs from local and summarized node attention are then merged to form the final output.

The CCA module projects the encoder output into a fixed length in each layer of the decoder to reduce complexity efficiently. Because the projection is performed separately with different weights in each layer, the information loss from linear projection is minimized and the decoder layer can capture variant features and information from the encoder output.
Our method successfully achieved a complexity of order $O(l)$ for a sequence length of $l$ while capturing locality and global information.

The CCA is used instead of encoder distillation as in Informer to reduce computational complexity in cross-attention while replacing the self-attention mechanism with GSA.

\begin{figure}[tbp]
 \centering
 \includegraphics[width=0.9\textwidth]{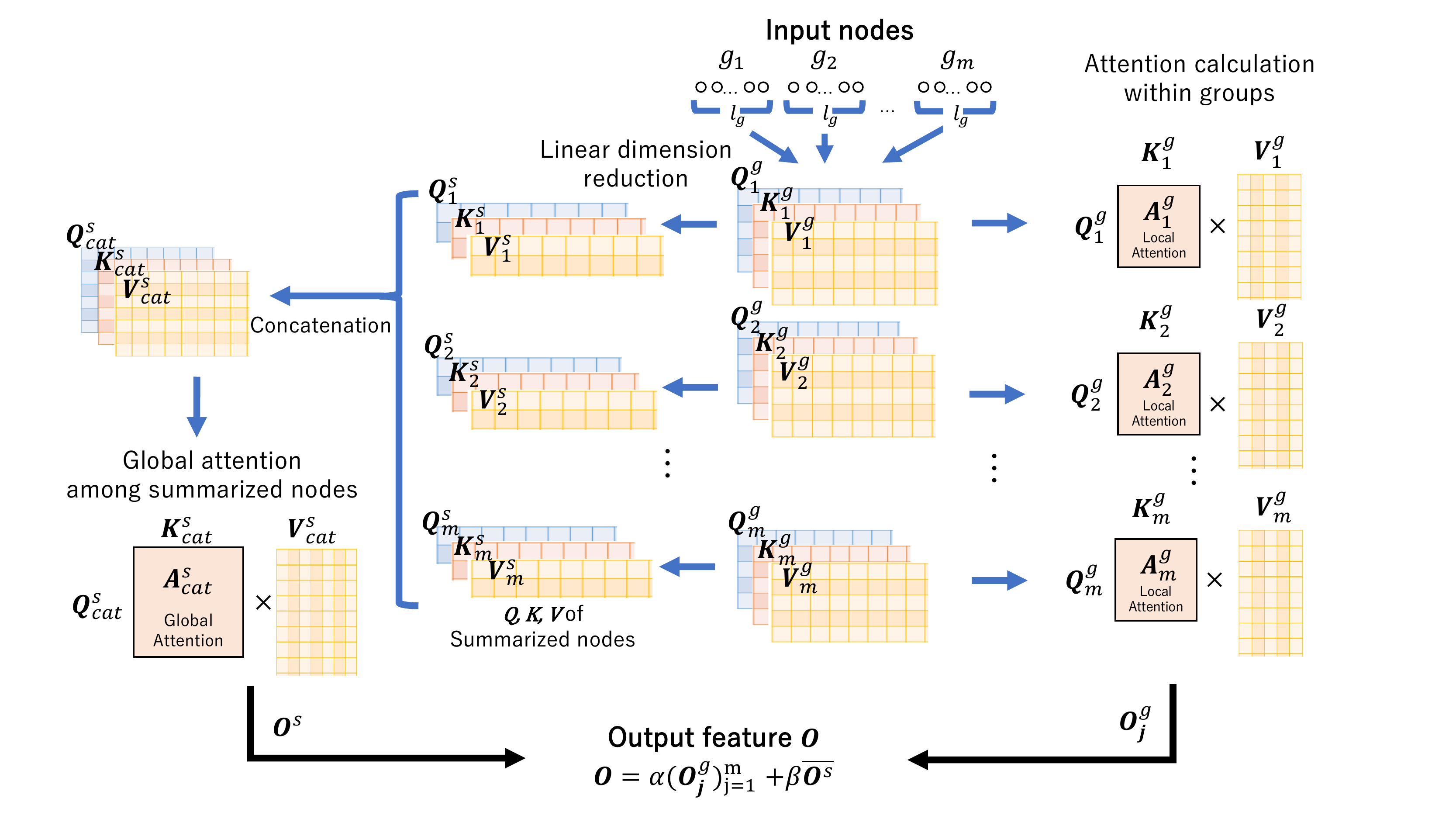}
 \caption[Grouped self-attention]{A detailed illustration of the Grouped Self-Attention module.}
 \label{fig:grouped_self_attention_detailed}
\end{figure}

\subsection{Grouped Self-Attention}
In this section, we provide a detailed explanation of the proposed Grouped Self-Attention (GSA) mechanism. Overall, we utilize the concept that self-attention reflects the interaction among the input nodes into outputs. The input nodes are divided into several groups to calculate the local attention within the groups, and summarized nodes generated from each groups are concatenated to compute another self-attention to reflect the information interaction among other groups.

For detailed explanation, first, for each input node $n_i\in \R^d$ ($i\in\{1, ..., l\}$), query $\vq_{i}\in \R^d$, key $\vk_{i}\in \R^d$, and value $\vv_{i}\in \R^d$ are generated by linear projection as in canonical self-attention methods. The input nodes are divided into several groups within which the local attention is computed to capture the locality. 

To compute the attention function efficiently, all queries, keys, and values are concatenated to form matrices $\mQ\in \R^{l\times d}$, $\mK\in \R^{l\times d}$ and $\mV\in \R^{l\times d}$. Before computing the attention matrix, as shown in Fig.\ref{fig:grouped_self_attention} and Fig.\ref{fig:grouped_self_attention_detailed}, the input nodes are divided into several groups with a length of $l_g$. We define these groups of nodes as G = $\{g_j\in \mathbb{R}^{l_g\times d}\}_{j=1}^m$, whereas $m$ is the number of divided groups. If $l$ cannot be divided by $l_g$, the zero-padded nodes are added to match the length $ml_g$. Queries, keys, and values are generated by linear projection for each group. For example, queries, keys, and values of group $g_j$ are defined as $\mQ_j^g\in \R^{l_g\times d}$, $\mK_j^g\in \R^{l_g\times d}$, and $\mV_j^g\in \R^{l_g\times d}$ respectively. The attention output of group $g_j$, $\mO_j^g\in \R^{l_g\times d}$, is computed as given below in equation \ref{eq:attention2}

\begin{equation}
    \label{eq:attention2}
        \mO_j^g = Attention(\mQ_j^g, \mK_j^g, \mV_j^g).
\end{equation}

After computing the local attention among these groups, summarized nodes defined as S = $\{S_j\in \R^{l_s\times d}\}_{j=1}^m$ are computed. The queries, keys, and values of summarized nodes $S_i$ for group $g_j$ are calculated by applying linear projection matrices $\mE_q,\mE_k,\mE_v\in \R^{l_s\times l_g}$ as given below in equation \ref{eq:projection1}. 

\begin{equation}
    \label{eq:projection1}
        \mQ_j^s = \mE_q\mQ_j^g, \\
        \mK_j^s = \mE_k\mK_j^g, \\
        \mV_j^s = \mE_v\mV_j^g. 
\end{equation}
where $\mQ_j^s, \mK_j^s, \mV_j^s\in \R^{l_s\times d}$.

Self-attention is again calculated only with these queries, keys, and values of summarized nodes to reflect the information of other groups. To compute the self-attention among summarized nodes, all $\mQ_j^s, \mK_j^s$, and $\mV_j^s$ are concatenated to form $\mQ^s_{cat}, \mK^s_{cat}, \mV^s_{cat}\in \R^{ml_s\times d}$.

These concatenated queries, keys, and values are used to compute the global self-attention output $\mO^s$ as illustrated in Fig.\ref{fig:grouped_self_attention} which represents the projected global features of all input nodes. $\mO^s\in \R^{ml_s\times d}$ of equation \ref{eq:attention3} reflects the global information by calculating attention with summarized queries and keys across the groups. 

\begin{equation}
    \label{eq:attention3}
        \mO^s = Attention(\mQ^s_{cat}, \mK^s_{cat}, \mV^s_{cat}). 
\end{equation}

$\mO^s$ is then divided into $m$ segments where each segment $\mO_j^s\in \R^{l_s\times d}$ represents the output that reflects global features with local information in group $g_j$. $\mO_j^s$ is pooled by average pooling to form $\overline{\mO_j^s}\in \R^{1\times d}$, where $\overline{\mO_j^s}=\sum_{k=1}^{l_s}(\mO_j^s)_k$, and it is added to the local attention output $\mO_j^g$ to form the final output of group $g_j$, $\widetilde{\mO_j^g}\in \R^{l_g\times d}$ according to equation \ref{eq:final_output}.

\begin{equation}
    \label{eq:final_output}
        \widetilde{\mO_j^g} = \alpha \mO_j^g + \beta \overline{\mO_j^s}.
\end{equation}

$\alpha$ and $\beta$ are learnable parameters for each group that determine the reflection ratio of global output into local attention. All $\widetilde{\mO_j^g}$ are concatenated to form final output of the GSA layer $\mO\in \R^{ml_g\times d}$. This process allows the local attention to consider the global features while capturing locality.

The computational complexity of our proposed approach increases with the number of dot-production calculations in the attention, which is given as $m\times l_g^2 + (m\times l_s)^2$. Because $m$ is the number of groups and $l_g$ is the number of nodes included in one group, $m\times l_g$ equals the sequence length $l$, and because $l_s$ is the length of summarized nodes that qualifies $l_s<<l_g<l$, the algorithm has a complexity of order $O(l_g\times l)$. $l_g$ is a fixed length representing the number of nodes included in a single group, and thus the final complexity order becomes linear with respect to sequence length $l$.

To summarize, the GSA module is a linear-order self-attention mechanism based on a local attention mechanism that considers global features by receiving information from interactions of summarized nodes.

\subsection{Compressed Cross-Attention}
The cross-attention module in the canonical Transformer reflects the encoded input features from encoder outputs to the decoder layer. The keys and values for cross-attention are generated from the encoder output, and the queries are generated from the decoder input. The attention is computed using those queries, keys, and values similar to equation \ref{eq:self-attention}. The length of queries $l_Q$, generated from encoder output is equal to that of the input sequence, and the lengths of keys and values $l_K$ generated from decoder input are equal to that of the prediction sequence. This module thus exhibits a quadratic computational and space complexity of order $O(l_Q\times l_K)$ when the values of $l_Q$ and $l_K$ are similar.

We propose a Compressed Cross-Attention (CCA) module that performs linear projection on encoder output across all decoder layers to reduce complexity while minimizing information loss during projection. The output of the encoder is compressed by linear projection to a fixed length $l_{comp}$ in each cross-attention module in the decoder with separate weights. Information loss is minimized by compressing the encoder outputs multiple times in each decoder layer because using separate weights across all layers can extract various different features from the encoder output. Moreover, the computational complexity is reduced to order $O(l_Q\times l_{comp})$, which is linear because $l_{comp}$ is a changeable hyperparameter of fixed length.

\section{Experiments}
In this section, we evaluate the performance, memory efficiency, and computational efficiency of our proposed method compared with those of other related methods on time-series datasets that require long sequences. The evaluations were conducted with four baselines utilizing other variations of self-attention modules including the ProbSparse attention from Informer (\cite{zhou2021informer}), low-rank approximation of attention matrix in Linformer (\cite{wang2020linformer}), Local1d attention from Image Transformer (\cite{parmar2018image}), and LSH attention from Reformer (\cite{kitaev2020reformer}). The hyperparameter settings for these methods are given in \ref{hyper_other}

The time-series datasets were obtained from \cite{zhou2021informer}, and three datasets are selected to perform the experiments. \textbf{ETT} (Electricity Transformer Temperature), which is a crucial indicator in long-term electrical power deployment, includes datasets with two years of information of ETTh1, ETTh2 recorded with one-hour frequency and an ETTm1 dataset with 15-minute frequency. \textbf{ECL} (Electricity Consumption Load) contains hourly electricity consumption for 2 years. Finally, the \textbf{Weather} dataset consisted of local climatological data collected over four years. All datasets included features with multiple dimensions for a single period. Target values were selected in each dataset as “oil temperature” for ETT, “MT\_320” for ECL, and “wet bulb” for Weather. Our division of the dataset into training, validation, and testing sets followed that of \cite{zhou2021informer}.

\subsection{Hyper-parameter Settings}
\label{hyper_other}
The hyperparameters that need to be set for the proposed model are listed as the group node length $l_g$, the number of layers in the encoder and decoder $e_l, and d_l$, the summarized node length $l_s$, and a fixed length $l_{comp}$ in CCA module. Other parameters common to the canonical Transformer model follow \cite{vaswani2017attention}, such as the dimension of features in the multi-head attention module. A grid search was conducted for hyperparameter settings, and values of 64, 90 for $l_g$, 1, 3 for $e_l$ and $d_l$, and 4, 8 for $l_s$ were obtained. The parameters selected according to model performance were 64 for $l_g$, 3 for $e_l$ and $d_l$, and 4 for $l_s$. The parameters $e_l$ and $d_l$ were shared with the baseline models, and thus same values were used for other networks. Finally, a fixed compressed length $l_{comp}$ in the CCA module is set to 256.

Informer, Linformer, Image Transformer, and Reformer models were used as baselines in the experiments. As we utilized frameworks and code from Informer, the default parameters from \cite{zhou2021informer} were used with the Informer architecture. The projected dimension in the low-rank approximation of Linformer was set to 256, as in \cite{wang2020linformer}. For local1d attention in Image Transformer, the block length for each local block was set to 64 as in $l_g$ in grouped attention for a fair comparison with the proposed method. The bucket sizes required in Reformer were set to 48, 42, 72, 42, 42, 60, and 60 for sequence lengths 96, 168, 288, 336, 672, 720, and 1440 respectively because half of the sequence length should be divisible by the bucket size.

\subsection{Experimental Setup}
As mentioned above, ETTh1, ETTh2, ETTm1, ECL, and Weather datasets were used to evaluate the performance of models. For each model, global time stamp embedding and generative style decoder structure were applied as in \cite{zhou2021informer} to focus on comparing the self-attention modules. One difference among the networks other than the self-attention module was that Informer used encoder distillation to reduce the computational complexity of calculating cross-attention, and the CCA module was applied to reduce the computational complexity of other methods, including the proposed approach. 

In the experiment with multivariate settings, we fed the multi-dimensional features as input to predict all the features, including not only the target feature but also other features as well. In contrast, in the experiments following univariate settings, we input the single-dimensional feature of the target value to predict target features using the output of the models. We focused on comparing the self-attention modules as much as possible, and we also compared the performance and computational complexity for all the variations of self-attention.

\begin{table}[t]
  \caption{MSE losses results from experiments with multivariate and univariate settings}
  \label{table:Multivariate_MSE_loss}
  \centering
    \resizebox{140mm}{!}{
    \begin{tabular}[t]{c|c|c|c|c|c||c|c|c|c|c} \hline
     &  \multicolumn{5}{c||}{Multivariate} &  \multicolumn{5}{c}{Univariate}
    \\ \hline
      \rule{0pt}{2ex} Data(seq\_len) & Grouped (Ours) & Informer & Linformer & Local1d & Reformer & Grouped (Ours) & Informer & Linformer & Local1d & Reformer 
      \\ \hline \hline
    \Centerstack{ \rule{0pt}{2ex} ETTh1(168) \\ ETTh1(336) \\ ETTh1(720) \\ ETTh1(1440) \rule[-0.6ex]{0pt}{0pt}}
      & \Centerstack{0.8531 \\ 1.1047 \\ \textbf{1.0624} \\ \textbf{0.8488}} & \Centerstack{ 1.0538 \\ 1.0841 \\ 1.3786 \\ 1.2791} & \Centerstack{0.9580 \\ 1.2129 \\ 1.3199 \\ 1.0565} & \Centerstack{0.8152 \\ \textbf{1.0253} \\ \textbf{1.0716} \\ 1.0797} & \Centerstack{\textbf{0.7699} \\ 1.1014 \\ 1.1505 \\ 1.0394} & \Centerstack{0.3844 \\ 0.3001 \\ \textbf{0.2062} \\ 0.4243} & \Centerstack{\textbf{0.1537} \\ 0.3031 \\ 0.2627 \\ \textbf{0.3732}} & \Centerstack{0.4034 \\ 0.4048 \\ 0.2542 \\ 0.4818} & \Centerstack{0.3204 \\ 0.4048 \\ 0.2587 \\ 0.5918} & \Centerstack{0.3274 \\ \textbf{0.2757} \\ 0.3505 \\ 0.7048} \\ \hline

      \Centerstack{ \rule{0pt}{2ex} ETTh2(168) \\ ETTh2(336) \\ ETTh2(720) \\ ETTh2(1440) \rule[-0.6ex]{0pt}{0pt}}
      & \Centerstack{9.9923 \\ 4.2793 \\ \textbf{2.7107} \\ 3.0034} & \Centerstack{10.9307 \\ 3.3196 \\ 4.4807 \\ 4.6793} & \Centerstack{\textbf{5.8715} \\ 5.0038 \\ 3.5199 \\ 2.9800} & \Centerstack{9.8326 \\ \textbf{2.8158} \\ 2.9131 \\ 2.5742} & \Centerstack{8.9058 \\ 3.0375 \\ 2.9069 \\ \textbf{2.5441}} & \Centerstack{\textbf{0.1844} \\ \textbf{0.2555} \\ 0.2737 \\ \textbf{0.2388}} & \Centerstack{0.2742 \\ 0.3220 \\ 0.2920 \\ 0.2616} & \Centerstack{0.3066 \\ \textbf{0.2515} \\ 0.2352 \\ \textbf{0.2440}} & \Centerstack{0.2330 \\ 0.2697 \\ \textbf{0.2220} \\ 0.2601} & \Centerstack{0.2611 \\ 0.3008 \\ 0.2619 \\ 0.2484} \\ \hline

      \Centerstack{ \rule{0pt}{2ex} ETTm1(96) \\ ETTm1(288) \\ ETTm1(672) \\ ETTm1(1440) \rule[-0.6ex]{0pt}{0pt}}
      & \Centerstack{\textbf{0.5649} \\ 0.9367 \\ \textbf{0.8121} \\ 1.0737} & \Centerstack{0.6190 \\ \textbf{0.8505} \\ 1.0566 \\ 1.1128} & \Centerstack{\textbf{0.5721} \\ 0.8644 \\ 0.8775 \\ 1.1834} & \Centerstack{0.5902 \\ 1.0439 \\ 0.9359 \\ \textbf{1.0105}} & \Centerstack{0.5813 \\ \textbf{0.8460} \\ 0.9602 \\ 1.1913} & \Centerstack{0.0883 \\ 0.3285 \\ 0.3663 \\ 0.2973} & \Centerstack{0.1011 \\ 0.3594 \\ \textbf{0.3457} \\ \textbf{0.2485}} & \Centerstack{\textbf{0.0511} \\ 0.3327 \\ 0.3948 \\ 0.3165} & \Centerstack{0.1622 \\ \textbf{0.2811} \\ 0.3943 \\ \textbf{0.2516}} & \Centerstack{0.1020 \\ 0.3642 \\ \textbf{0.3508} \\ 0.3689} \\ \hline

      \Centerstack{ \rule{0pt}{2ex} ECL(168) \\ ECL(336) \\ ECL(720) \\ ECL(1440) \rule[-0.6ex]{0pt}{0pt}}
      & \Centerstack{0.2785 \\ 0.3122 \\ \textbf{0.2850} \\ \textbf{0.2894}} & \Centerstack{0.2772 \\ 0.3023 \\ 0.3089 \\ 0.3134} & \Centerstack{\textbf{0.2619} \\ 0.3250 \\ \textbf{0.2775} \\ \textbf{0.2915}} & \Centerstack{0.2893 \\ \textbf{0.2944} \\ 0.2971 \\ 0.2948} & \Centerstack{0.2842 \\ 0.2988 \\ 0.2895 \\ 0.3071} & \Centerstack{\textbf{0.3371} \\ 0.3877 \\ 0.4504 \\ 0.4474} & \Centerstack{\textbf{0.3356} \\ 0.4073 \\ 0.4440 \\ 0.4704} & \Centerstack{0.3587 \\ \textbf{0.3698} \\ \textbf{0.3524} \\ 0.5003} & \Centerstack{0.3643 \\ 0.4426 \\ 0.5085 \\ \textbf{0.3052}} & \Centerstack{0.3700 \\ 0.4059 \\ 0.4287 \\ 0.4462} \\ \hline

      \Centerstack{ \rule{0pt}{2ex} WTH(168) \\ WTH(336) \\ WTH(720) \\ WTH(1440) \rule[-0.6ex]{0pt}{0pt}}
      & \Centerstack{0.5848 \\ 0.6263 \\ \textbf{0.6088} \\ \textbf{0.5769}} & \Centerstack{0.6246 \\ 0.6273 \\ 0.6609 \\ 0.6343} & \Centerstack{0.5986 \\ \textbf{0.6155} \\ 0.6329 \\ 0.6147} & \Centerstack{\textbf{0.5685} \\ 0.6273 \\ 0.6516 \\ 0.5983} & \Centerstack{0.6345 \\ 0.6414 \\ 0.6732 \\ 0.6114} & \Centerstack{0.2149 \\ 0.2755 \\ 0.2830 \\ \textbf{0.2190}} & \Centerstack{0.2587 \\ 0.2658 \\ 0.3532 \\ 0.2475} & \Centerstack{\textbf{0.2032} \\ 0.2635 \\ 0.2401 \\ 0.2251} & \Centerstack{0.2615 \\ 0.2539 \\ 0.2572 \\ 0.2357} & \Centerstack{0.2160 \\ \textbf{0.2411} \\ \textbf{0.2250} \\ \textbf{0.2150}} \\ \hline
      \rule{0pt}{2ex} count & 9 & 1 & 6 & 6 & 3 & 6 & 5 & 6 & 4 & 5 \\ \hline
      
    \end{tabular}
    }
\end{table}

\subsubsection{Performance Evaluation}
The results are shown in Table \ref{table:Multivariate_MSE_loss}, which indicates the results of the experiments with multivariate and univariate settings. For each setting, the best-performing value is shown in bold, and the difference between the second-best value is shown in bold as well if it was under $0.01$. We used the mean square error (MSE) loss as an evaluation metric. MSE loss was also used as a loss function to train all the models and the learning rate was set to $0.0001$. The sequence lengths varied from 168 to 1440, as shown in the seq\_len section of Table \ref{table:Multivariate_MSE_loss}. ETTm1 was used with a slightly different seq\_len range for comparison with the results reported in \cite{zhou2021informer}. The variable seq\_len represents the input sequence length and prediction sequence length, and the length of labels used in generative style decoder input was set to half of the seq\_len. The sequence lengths under 64 were removed and compared to the results shown in \cite{zhou2021informer} because the group node length $l_g$ was selected as 64, and if the input sequence length was under 64, it behaved equivalently with self-attention in canonical transformer. 

Table \ref{table:Multivariate_MSE_loss} illustrates that the performance of Grouped Transformer was the best across the experiments, especially for the experiment with multivariate settings. 

Baselines other than Informer performed better compared to the experimental results in \cite{zhou2021informer}. However, in our experimental settings, all the baselines shared the global time stamp embedding and generative style decoder structure proposed in \cite{zhou2021informer}. Moreover, our proposed method, Linformer, Local1d, and Reformer utilized the CCA mechanism to reduce the computational cost. Because all the methods shared the Informer framework including global time stamp embedding and generative style decoder, the only difference was the mechanism of self-attention, and thus the baselines other than Informer performed better as well.

\begin{figure}[tbp]
 \centering
 \includegraphics[width=0.9\textwidth]{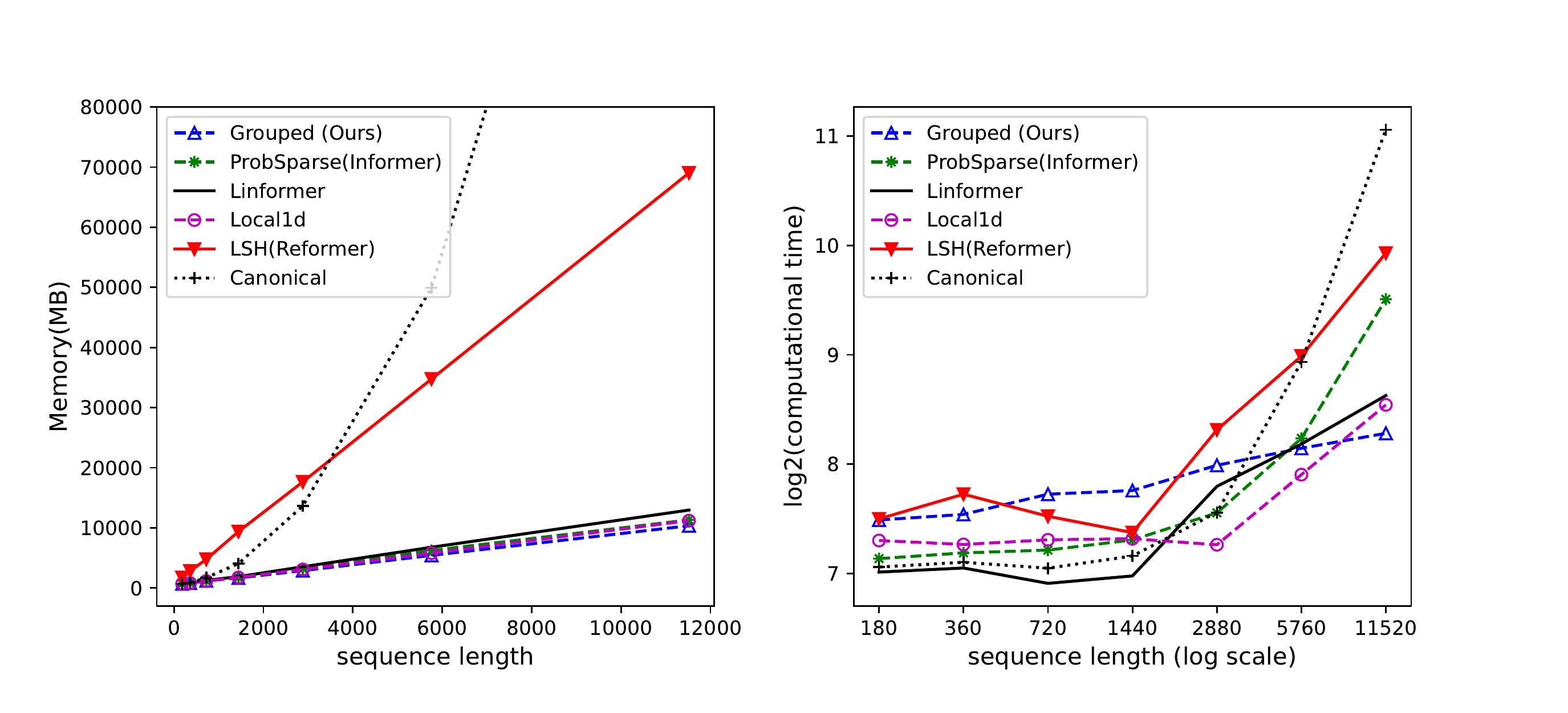}
 \caption[Memory usage]{The figure at left shows the memory usage of the attention mechanisms with respect to the sequence length. The memory usage of canonical Transformer in sequence length of 11520 was 190341MB. The figure on the right illustrates the average computational time per 1000 iterations of attention mechanisms with respect to sequence length on a log scale for easier comparison.}
 \label{fig:memory_usage}
\end{figure}

\subsubsection{Memory and Computational Complexity Evaluation}
We conducted experiments to evaluate the computational complexity of our proposed approach and each baseline to analyze the memory usage and computational speed according to the sequence length fed to the self-attention layers. This experiment was conducted for every model with a variation of sequence lengths of 180, 360, 720, 1440, 2880, 5760, and 11520 using the Weather dataset. These sequence lengths indicate the input and prediction sequence lengths. Because the purpose was to check the order of memory usage and computational speed of self-attention modules, the other parameters were set to small values to exclude any potential confounding effects. $e_l$ and $d_l$ were set to 3 as in the experiment setup, and the dimension of features $d$ was 256 with four-headed attention, the hidden dimension in the feed-forward network between attention layers was 256, and the length of labels used in the generative style decoder input was 0. Moreover, the Weather dataset was used with a multivariate setting. Memory usage was calculated during the training and backpropagation, and the computational time per iteration was calculated as the average consumed time per iteration for one epoch of training. This experiment is conducted on Quadro RTX 8000 GPUs. Fig.\ref{fig:memory_usage} shows the results of the experiments on memory usage computational time.

As shown in Fig.\ref{fig:memory_usage}, the proposed method efficiently minimized memory usage and computational cost. Our proposed method exhibited the lowest value in terms of both memory usage and computational time, especially with a longer sequence length such as 11520. While utilizing the least computing resources, our model achieved performance comparable to or better than that of the baseline methods.

\subsubsection{Ablation Study}
We conducted an ablation study on the GSA module to demonstrate the effect of reflecting global information on local attention. Table \ref{table:ablation_global} shows the MSE results of the ablation study. The experiment was conducted in a multivariate setting. Grouped$^{\dag}$ in Table \ref{table:ablation_global} only utilized $\mO^g_j$ instead of the final output $\widetilde{\mO_j^g}$ calculated in equation \ref{eq:final_output} to exclude the effect of global attention. To verify the premise that reflecting the global information would be beneficial to process the input with a long sequence length and to capture long-range dependencies, the ablation study was conducted with sequence lengths over 288.

The results of the ablation study shown in Table \ref{table:ablation_global} indicate that global feature reflection by summarized nodes in GSA was effective to capture long-range dependencies, especially for all the settings with seq\_len=1440, Grouped$^{\dag}$ did not perform as well as the proposed method.

\begin{table}[t]
  \caption{MSE results of ablation study on global attention reflection in GSA module with multivariate setting}
  \label{table:ablation_global}
  \centering
  
    \resizebox{100mm}{!}{
    \begin{tabular}[t]{|c|c|c|}\hline
      \rule{0pt}{2ex} Data(seq\_len) & Grouped & Grouped$^{\dag}$ \\
         \hline \hline
    \Centerstack{ \rule{0pt}{2ex} ETTh1(336) \\ ETTh1(720) \\ ETTh1(1440) \rule[-0.6ex]{0pt}{0pt}}
      & \Centerstack{\textbf{1.1047} \\ \textbf{1.0624} \\ \textbf{0.8488}} & \Centerstack{1.2464 \\ 1.1576 \\ 1.1332} \\ \hline

      \Centerstack{ \rule{0pt}{2ex} ETTh2(336) \\ ETTh2(720) \\ ETTh2(1440) \rule[-0.6ex]{0pt}{0pt}}
      & \Centerstack{4.2793 \\ 2.7107 \\ \textbf{3.0034}} & \Centerstack{\textbf{3.6806} \\ \textbf{2.5103} \\ 3.2622} \\ \hline

      \Centerstack{ \rule{0pt}{2ex} ETTm1(288) \\ ETTm1(672) \\ ETTm1(1440) \rule[-0.6ex]{0pt}{0pt}}
      & \Centerstack{0.9367 \\ \textbf{0.8121} \\ \textbf{1.0737}} & \Centerstack{\textbf{0.9006} \\ 0.9268 \\ 1.1138} \\ \hline
    \end{tabular}
    \hfill
    \begin{tabular}[t]{c|c|c|}  \hline
      \rule{0pt}{2ex} Data(seq\_len) & Grouped & Grouped$^{\dag}$ \\
      \hline \hline

      \Centerstack{ \rule{0pt}{2ex} ECL(336) \\ ECL(720) \\ ECL(1440) \rule[-0.6ex]{0pt}{0pt}}
      & \Centerstack{0.3122 \\ \textbf{0.2850} \\ \textbf{0.2894}} & \Centerstack{\textbf{0.3009} \\ 0.2980 \\ 0.2948} \\ \hline

      \Centerstack{ \rule{0pt}{2ex} WTH(336) \\ WTH(720) \\ WTH(1440) \rule[-0.6ex]{0pt}{0pt}}
      & \Centerstack{0.6263 \\ \textbf{0.6088} \\ \textbf{0.5769}} & \Centerstack{\textbf{0.6240} \\ 0.6278 \\ 0.5937} \\ \hline
      
    \end{tabular}
    }
\end{table}

\section{Conclusion}
In this study, we have proposed a GSA module designed to capture locality with local attention while reflecting global features by calculating attention among summarized nodes. We also have proposed a CCA module to reduce the computational cost of computing the cross-attention. Our method successfully reduced computational complexity and memory usage and achieved performance comparable to or better than existing methods with time-series data. Our experimental results demonstrate that the proposed approach can capture long-range dependencies with an efficient attention mechanism in the Transformer model.

For future work, we would like to apply our proposed modules to other tasks that require long sequences such as image pixel generation or protein sequence modeling tasks. Moreover, because each module can be applied independently, GSA module can be applied to encoder-only structure of Transformer such as BERT \cite{devlin2018bert} to testify the performance and enlarge the maximum length of sequence input with its memory-efficiency.

\subsubsection*{Acknowledgments}
This work was partially supported by JST AIP Acceleration Research JPMJCR20U3, Moonshot R\&D Grant Number JPMJPS2011, CREST Grant Number JPMJCR2015, JSPS KAKENHI Grant Number JP19H01115 and Basic Research Grant (Super AI) of Institute for AI and Beyond of the University of Tokyo. We also thank Hanqin Wang for helpful discussions and reviewing the manuscript.

\bibliography{iclr2023_conference}
\bibliographystyle{iclr2023_conference}


\end{document}